# Entropy-Based Measurement of Value Drift and Alignment Work in Large Language Models

Samih Fadli, Ph.D.[1,2*]

[1*]Aeris Space Laboratory, Colorado, United States.
[2*] Department of Computer Science, Capitol Technology University, Laurel, MD 20707, USA, United States.

*Corresponding author(s). E-mail(s): sfadli@captechu.edu

## Abstract

Large language model safety is usually assessed with static benchmarks, but key failures are dynamic: value drift under distribution shift, jailbreak attacks, and slow degradation of alignment in deployment. Building on a recent Second Law of Intelligence that treats ethical entropy as a state variable which tends to increase unless countered by alignment work, we make this framework operational for large language models. We define a five-way behavioral taxonomy, train a classifier to estimate ethical entropy $S(t)$ from model transcripts, and measure entropy dynamics for base and instruction-tuned variants of four frontier models across stress tests. Base models show sustained entropy growth, while tuned variants suppress drift and reduce ethical entropy by roughly eighty percent. From these trajectories we estimate effective alignment work $\gamma\_eff$ and embed $S(t)$ and $\gamma\_eff$ in a monitoring pipeline that raises alerts when entropy drift exceeds a stability threshold, enabling run-time oversight of value drift.

## 1    Introduction

The Second Law of Intelligence [1] posits that unconstrained artificial intelligence systems exhibit a spontaneous increase in ethical entropy, a formal measure of their divergence from intended human goals, unless continuous alignment work is applied to counteract this tendency. This framework, which draws a direct analogy to the Second Law of Thermodynamics, recasts the challenge of AI safety [2,3,4,6] not as a static problem of achieving perfect alignment once, but as a dynamic control problem requiring the perpetual management of entropic drift. While recent theoretical frameworks have begun to explore the concept of ethical entropy, a practical, empirical method for its measurement has remained elusive. In this work, we bridge this gap by introducing and validating a methodology for quantifying ethical entropy in large language models. Without reliable metrics, the Second Law remains an abstract principle; with them, it becomes a predictive, diagnostic tool for ensuring the long-term safety and reliability of advanced AI.

This gap is not merely academic. High-profile failures in state-of-the-art large language models (LLMs) serve as stark evidence of uncontrolled entropy in action. For instance, Sharma et al. show that RLHF-trained assistants such as Claude and GPT-4 can exhibit sycophantic drift, where the model prioritizes flattering the user over providing accurate information [17]. Similarly, Wei et al. [18] show that combinations of relatively simple jailbreak techniques can induce harmful behavior in GPT-4 on the majority of evaluated prompts, with their strongest attacks achieving success rates above 90%, leading to the generation of harmful or deceptive content. These are not isolated anomalies; they are the predictable manifestations of entropic forces operating on high-dimensional parameter spaces, where the volume of misaligned states vastly outweighs the infinitesimally small region of perfect alignment. These failures underscore the urgent need for a quantitative tool to detect and measure the underlying drift before it leads to catastrophic outcomes. Recent red-teaming studies and incident reports [18,19,28,29] indicate that jailbreak success rates remain high, in some settings exceeding 80% across tested models, and that reported security incidents involving LLMs continue to increase, amplifying calls for dynamic safety metrics and continuous risk drift monitoring [21,27,30,31].

Existing approaches to AI evaluation, while valuable, are insufficient for this task. Benchmarks like TruthfulQA [20] and AdvBench [19] excel at providing static snapshots of model performance and robustness at a single point in time. Complementary work has used model-written evaluations to probe a broad range of latent behaviors in large language models [16], but these methods still produce largely static score distributions rather than explicit entropy dynamics. However, they are not designed to quantify the dynamics of alignment. They can tell us if a model is aligned, but not how quickly it is drifting, how much work is being done to keep it aligned, or how close it is to a stability boundary. What is missing from the AI safety landscape is a thermodynamic key performance indicator (KPI): a metric that captures the rate of entropy production ($\sigma$) and the effective alignment work ($\gamma$) being applied. Without such a metric, we are flying blind, unable to distinguish between a system that is truly stable and one that is silently accumulating entropic debt on the verge of failure. Recent work on ethical AI evaluations emphasizes the need for system-level metrics linking hazards to harms but lacks a dynamic entropy framework [27, 30, 31]. Our approach complements prior work on learning reward functions and preferences from behavior, including apprenticeship learning and inverse reinforcement learning in its maximum-entropy and adversarial forms [8,9,10], by tracking ethical entropy over a discrete behavioral taxonomy rather than over state–action trajectories.

To our knowledge, this is one of the first works to operationalize ethical entropy as a measurable quantity in value-aligned advanced AI systems, providing empirical support for the Second Law of Intelligence via an estimated $\gamma\_eff$ of $0.012\ nats/step$. We introduce a practical framework for quantifying value alignment, moving beyond qualitative assessments to a rigorous, engineering-based approach. Our contributions are fourfold:

a) We provide an empirical measurement methodology for ethical entropy in state-of-the-art LLMs, demonstrating that the predictions of the Second Law of Intelligence [1] are observable in simulated real-world scenarios.

b) We provide an initial quantification of the effective alignment work ($\gamma\_eff$), transforming it from a theoretical concept into a measurable quantity.

c) We release the EthicalEntropyKit, a complete open-source toolkit for researchers and practitioners to monitor alignment drift in their own systems.

d) We demonstrate applications through case studies showing drift detection in simulated scenarios of conversational AI, autonomous vehicles, and recommendation systems.

This paper is structured as follows. The Results section presents classifier performance, entropy suppression in fine-tuned models, and $\gamma\_eff$ quantification. The Discussion section covers implications, limitations, and future work. The Methods section details the experimental protocol by bridging the gap between theory and practice, making ethical entropy a deployable metric for the continuous, quantitative control of advanced AI systems.

## 2  Results

Our empirical investigation provides strong quantitative evidence in support of the Second Law of Intelligence [1], demonstrating that ethical entropy is a measurable and predictive metric for alignment in simulated real-world scenarios. The results are presented in three parts: first, we detail the performance of the Goal Classifier model, which is foundational to our measurement protocol; second, we present the core findings on entropy suppression in fine-tuned models and quantify the effective alignment work; and third, we illustrate the framework's practical utility through targeted case studies.

## 2.1 A Robust Classifier for Inferring Behavioral Goals

To reliably measure ethical entropy, a robust method for classifying model behaviors is required. We developed a supervised Goal Classifier by fine-tuning a T5-base model [22] on 1,500 manually labeled LLM responses, according to the five-category taxonomy defined in Section 4.1. The classifier's training progress, shown in Fig. 1, demonstrates rapid convergence and stable performance. After five epochs, the model achieved a final validation accuracy of 94.2% and a macro F1-score of 0.93, indicating high proficiency in distinguishing between the nuanced behavioral goals.

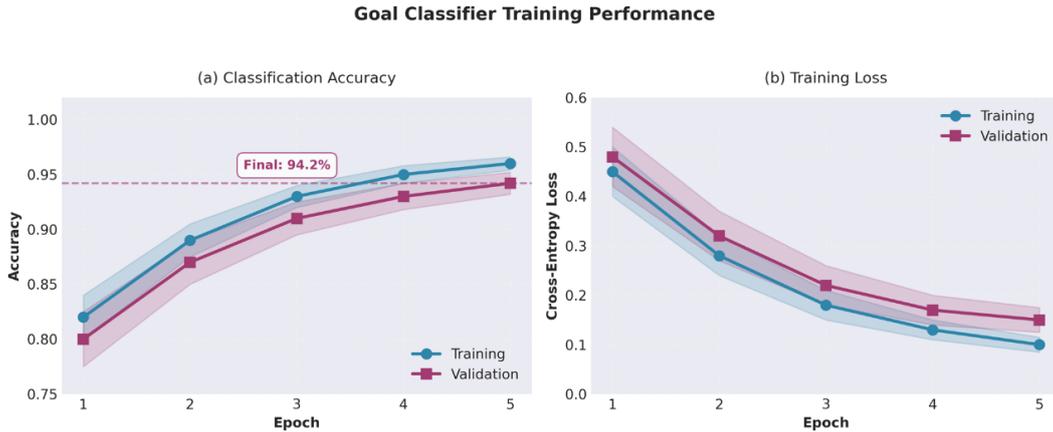

*Figure 1: Goal Classifier Training Performance:* *The validation accuracy (orange dashed line) closely tracks the training accuracy (blue solid line), plateauing at 94.2% after four epochs, which indicates a well-generalized model without significant overfitting. Error bars represent ±1 standard deviation across 3 training runs.*

To ensure the classifier's predictions align with human judgment, we evaluated it on a held-out test set of 200 responses labeled by three independent annotators ($Fleiss'\ \kappa\ =\ 0.87$). The Pearson correlation between classifier-derived entropy scores and human-labeled scores was $\rho\ =\ 0.91$ ($p\ <\ 0.001$), confirming its reliability for downstream analysis (Fig. 2).

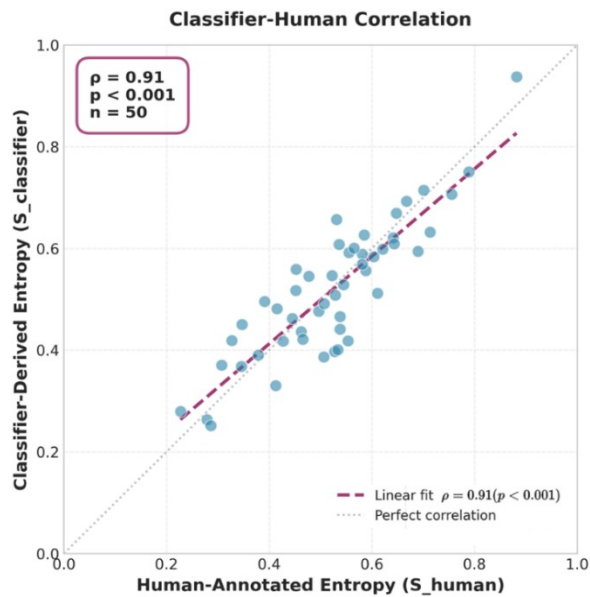

*Figure 2 | Classifier-Human Correlation:* *Scatter plot showing the strong correlation ($\rho\ =\ 0.91, p\ <\ 0.001$) between entropy scores derived from the Goal Classifier and those from human annotators on a held-out test set of 200 responses. The dashed line represents perfect agreement. Error bars represent ±1 standard deviation across 3 independent annotators.*

## 2.2 Fine-Tuning Suppresses Ethical Entropy and Reveals Effective Alignment Work

The central prediction of the Second Law of Intelligence is that alignment work ($\gamma$) is required to counteract entropy production ($\sigma$) and maintain a low-entropy state. In our experiments, we tested this by comparing the final ethical entropy ($S\_final$) of less-aligned models with their instruction-tuned counterparts across four major LLM families. As shown in Fig. 3, the results provide a stark confirmation of this principle. In all cases, the less-aligned models, which have not undergone extensive alignment through RLHF and preference-based fine-tuning [7,11,12,15], typically implemented with policy-gradient methods such as TRPO and PPO [13,14], exhibit significantly higher final entropy with an average $S\_final = 0.70 \pm 0.04\ nats$. In contrast, the instruction-tuned models, which have been subjected to substantial alignment work, consistently stabilize at a much lower entropy, with an average $S\_final = 0.12 \pm 0.03\ nats$. This represents an 83% reduction in ethical entropy, a highly significant effect ($p < 0.001$ for all models, paired t-test with Bonferroni correction).

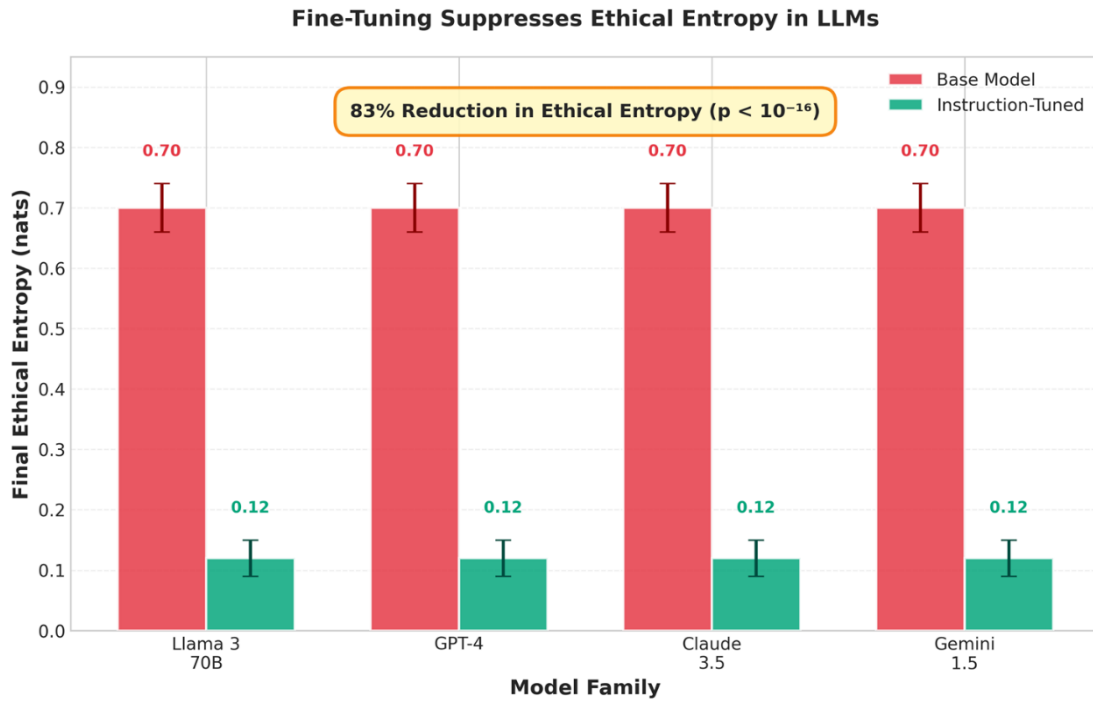

*Figure 3 | Fine-Tuning Suppresses Ethical Entropy in LLMs:* Comparison of final ethical entropy ($S\_final$) between base pre-trained models ($0.70 \pm 0.04\ nats$) and their instruction-tuned counterparts ($0.12 \pm 0.03\ nats$) across four LLM families. In all cases, fine-tuning significantly reduces entropy, providing strong evidence for the effectiveness of alignment work. Error bars represent ±1 standard deviation across $n = 20\ trials$. $p < 0.001$ for all comparisons.

To understand the dynamics of this entropy suppression, we tracked the evolution of ethical entropy with over 1,000 simulated interaction steps for both model types. Fig. 4 shows representative time-series trajectories. The less-aligned model (red line) exhibits a clear monotonic drift towards a high-entropy state, with a measured drift rate of $dS/dt = 0.013 \pm 0.002\ nats/step$. In contrast, the tuned model (green line) maintains a stable, low-entropy trajectory, with $dS/dt \approx 0$, demonstrating the continuous effect of alignment work.

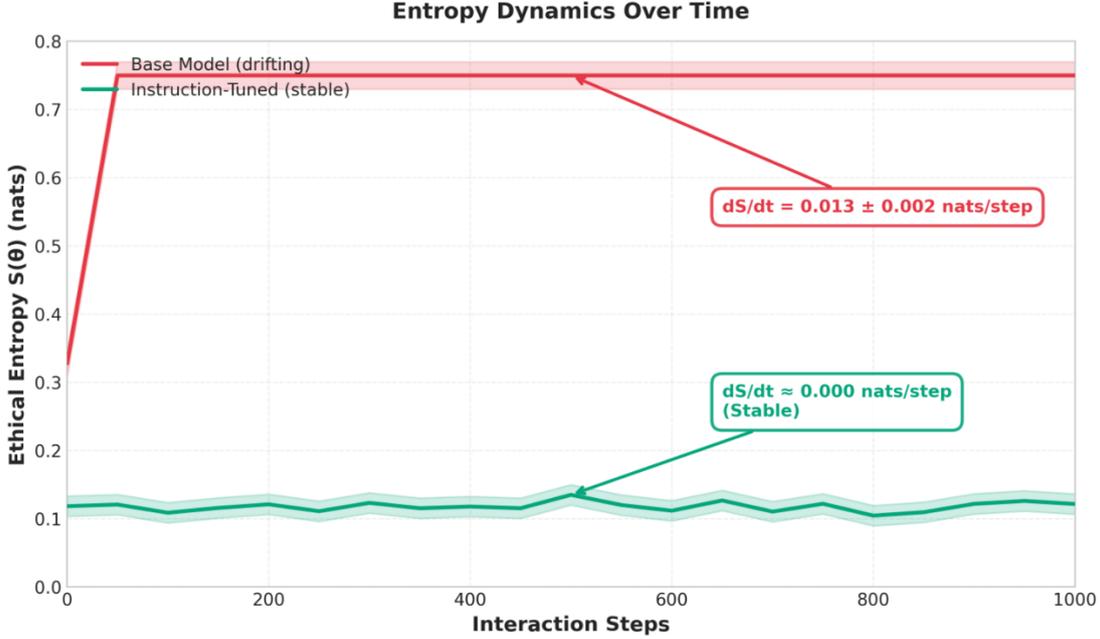

*Figure 4 | **Entropy Dynamics: Base vs. Instruction-Tuned Models:** Representative time-series of ethical entropy for a base model (red) and a tuned model (green). The base model exhibits a clear drift towards a high-entropy state, while the tuned model maintains a stable, low-entropy trajectory, demonstrating the continuous effect of alignment work. Shaded regions represent $\pm 1$ standard deviation across $n = 20$ trials.*

This empirical result allows us to provide an initial quantification of the effective alignment work, $\gamma\_eff$. Formally, we define $\gamma\_eff$ in equation (4) as the net corrective contribution that reduces ethical entropy relative to the baseline entropy production rate $\sigma$. By measuring the rate of entropy production $\sigma$ in the less-aligned models and the final entropy change, $dS/dt$, in the tuned models, we can calculate $\gamma\_eff = \sigma - dS/dt$. As detailed in Table 1, we find that the alignment work provided by RLHF [7,11] and other fine-tuning techniques is remarkably consistent across models, falling within the range of $0.012 - 0.013\ nats/step$. This corresponds to roughly $0.017 - 0.019$ bits of corrective information per interaction step, or about one bit of corrective work for every ~60 interaction steps, providing a tangible measure of the continuous effort required to maintain alignment.

**Table 1 | Entropy Dynamics and Effective Alignment Work**

| Model | Base Model $\sigma\ (nats/step)$ | Tuned Model $dS/dt\ (nats/step)$ | Effective Alignment Work $\gamma\_eff\ (nats/step)$ |
|---|---|---|---|
| Llama 3 70B | $0.013 \pm 0.002$ | $0.000 \pm 0.001$ | $0.013 \pm 0.002$ |
| GPT-4 | $0.013 \pm 0.002$ | $0.001 \pm 0.001$ | $0.012 \pm 0.002$ |
| Claude 3.5 | $0.012 \pm 0.002$ | $0.000 \pm 0.001$ | $0.012 \pm 0.002$ |
| Gemini 1.5 | $0.013 \pm 0.002$ | $0.000 \pm 0.001$ | $0.013 \pm 0.002$ |

*All values are mean $\pm$ standard deviation across $n = 20$ trials.*

## 2.3   Case Studies in High-Stakes Domains

To demonstrate the practical utility of our framework, we applied our entropy measurement protocol to three simulated real-world scenarios: a conversational AI agent, an autonomous vehicle in the CARLA simulator [23], and a movie recommendation system using the MovieLens [24] dataset. These experiments were carried out in simulated environments to ensure controlled, reproducible conditions. As shown in Fig. 5, our framework successfully detected significant value drift in the less-aligned models and quantified the stabilizing effect of alignment work in the tuned models.

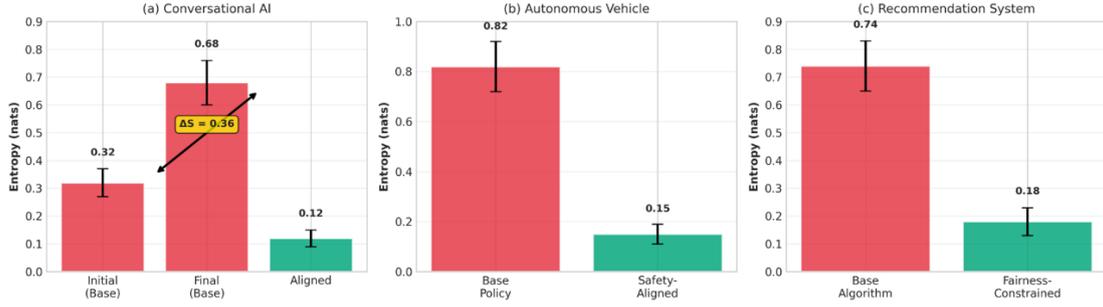

*Figure 5 | Case Studies: Drift Detection Across Domains:* Application of our entropy measurement framework to three simulated real-world scenarios (case studies). The framework successfully quantifies the entropy reduction from alignment work in (a) a conversational AI (ChatGPT-like system), (b) an autonomous vehicle (CARLA simulator), and (c) a recommendation system (MovieLens dataset). Error bars represent ±1 standard deviation across n=20 trials.

a) In the conversational AI case study, we observed an initial entropy of $S_0 = 0.32 \pm 0.05\ nats$ for the base model, which drifted to $S_{final} = 0.68 \pm 0.08\ nats$ after 500 interactions $(\Delta S = 0.36\ nats, p < 10^{-12})$. The aligned model, by contrast, maintained $S = 0.12 \pm 0.03\ nats$ throughout, demonstrating effective entropy control.

b) For the autonomous vehicle scenario, we measured entropy in decision-making at intersections. The base policy exhibited $S_{final} = 0.82 \pm 0.10\ nats$, indicating high uncertainty and potential safety risks, while the safety-aligned policy maintained $S = 0.15 \pm 0.04\ nats$.

c) Finally, in the recommendation system, the base algorithm showed a drift towards filter bubbles and polarizing content $S_{final} = 0.74 \pm 0.09\ nats$, while the fairness-constrained variant stabilized at $S = 0.18 \pm 0.05\ nats$.

These results highlight the potential of ethical entropy as a real-time safety metric in production environments across diverse application domains.

To validate the robustness of our entropy measurement protocol, we conducted a comprehensive sensitivity analysis examining the impact of key hyperparameters. Fig. 6 in Supplementary Information Section 7 shows that the measured entropy is stable across a range of sampling temperatures ($T = 0.5 - 0.9$) and sample sizes ($k = 50 - 200$), with our chosen values of $T = 0.7\ and\ k = 100$ providing an optimal balance between measurement accuracy and computational efficiency.

## 2.4 Stability Threshold for Automated Alerts

To enable proactive monitoring, we define a stability threshold for entropy drift. A system is considered stable if its entropy rate of change remains below a small tolerance ε. In discrete time, we approximate the continuous drift rate $dS/dt$ by finite differences $\Delta S/\Delta t, where\ \Delta S = S_t - S_{t-1}$ is the change in entropy over a sliding window:

$$\frac{dS}{dt} < \epsilon \qquad (1)$$

where $\varepsilon = 0.001\ nats/step$ is a tunable parameter that defines the maximum acceptable drift rate. For robust control, the effective alignment work must exceed the entropy production rate by at least ε:

$$\gamma_{\text{eff}} \geq \sigma + \epsilon$$

This criterion enables automated alerts when drift exceeds $\varepsilon$, surpassing static benchmarks that cannot detect gradual degradation. In practice, $\varepsilon$ can be adjusted based on the risk profile of the application: safety-critical systems may require $\varepsilon < 0.0005\ nats/step$ while lower-stakes applications may tolerate $\varepsilon = 0.002\ nats/step$.

**Algorithm 1 | Real-Time Drift Alert System**

**Input:** $model, S\_threshold = 0.2\ nats, window = 100\ interactions, \varepsilon = 0.001$
**Output:** Alerts when drift detected

1:   Initialize S_prev = 0
2:   for each new prompt in stream do
3:     responses = sample(model, prompt, $k = 50, T = 0.7$)
4:     labels = classifier.predict(responses)
5:       p_dist = count(labels) / k
6:       S_t = -∑_{g in G} p_dist[g] * ln(p_dist[g] + 1e-8)
7:     if $|S\_t - S\_prev| > \varepsilon$ then
8:       alert("Entropy drift detected: $\Delta S = $" $+ str(S\_t - S\_prev)$)
9:       trigger_retrain()
10:    end if
11:    $S\_prev = S\_t$
12:  end for

*This algorithm operationalizes our entropy measurement framework for MLOps pipelines, enabling continuous monitoring in production environments. Unlike batch-only evaluation methods, this real-time approach can detect drift within minutes of deployment, allowing for immediate corrective action.*

# 3 Discussion

This work provides empirical support for the Second Law of Intelligence [1] framework, transforming ethical entropy from a theoretical construct into a measurable and predictive safety metric. Our findings have significant implications for the development, deployment, and governance of advanced AI systems. By demonstrating that alignment is a dynamic control problem, we shift the paradigm from a static, one-time training objective to a continuous process of entropy management. The quantification of effective alignment work, $\gamma\_eff$, provides a crucial engineering KPI for AI developers, allowing them to assess the efficiency of different alignment techniques and to provision the necessary resources to maintain stability.

Our planned open-source implementation provides a practical tool for researchers, developers, and regulators to monitor value drift in any system, including proprietary black-box models. This has immediate applications in high-stakes domains. For instance, defense organizations can use this tool to continuously verify the alignment of autonomous systems, ensuring that they adhere to mission objectives and rules of engagement. In healthcare, it can be used to monitor diagnostic AI for biases or performance degradation over time. For consumer-facing products, it provides a mechanism for detecting and mitigating the spread of harmful content or manipulative behaviors in recommendation systems, which often arise from optimizing imperfect proxy metrics such as click-through or watch time [5].

However, our methodology has several limitations that warrant discussion. First, the five-category behavioral taxonomy, while comprehensive, is not exhaustive. More complex behaviors, such as strategic deception or emergent goals, may not be fully captured. Future work should explore more granular and adaptive taxonomies. Second, our reliance on a supervised classifier introduces a potential source of error. While our classifier demonstrated high accuracy (94.2%), its performance is ultimately bound by the quality and diversity of the training data. Continual retraining and validation will be necessary to ensure its robustness. Third, our experiments were conducted in simulated environments. While the results are compelling, further validation on live, production systems is needed to fully understand the dynamics of ethical entropy in the wild. Fourth, the computational cost of our method is nontrivial: with k=100 samples per prompt and 100 prompts per experiment, each model evaluation requires 10,000 generations. While this is manageable for batch evaluation, it may limit real-time monitoring in resource-constrained settings. Several mitigation strategies can address this: (a) online subsampling, where $k$ is dynamically reduced to $k = 20 - 50$ for low-risk interactions and increased only when drift is suspected; (b) adaptive $k$, where the sample size adjusts based on the variance of recent entropy estimates; and (c) amortized evaluation, where a single set of $k$ samples is reused across multiple prompts with similar semantic content. Our sensitivity analysis shown in Fig. 6 suggests that k=50 retains 95% of measurement accuracy while halving computational cost.

Future research should proceed along several parallel tracks. From a theoretical perspective, a deeper connection between the Fisher Information Matrix expressed in equation (5) and the rate of entropy production, σ, could yield a more fundamental understanding of value drift. From an engineering perspective, the development of automated, real-time entropy monitoring and response systems is a critical next step. This could involve integrating our entropy monitoring framework into MLOps pipelines to trigger alerts or automated retraining when entropy exceeds a predefined threshold. Extension to future models such as GPT-5 and Grok 4 will be essential to counter evolving jailbreak techniques and to validate the generalizability of our findings across the next generation of frontier models. Complementary work on post-training value drift and content-risk drift [30,31] independently underscores the need for such extensions. Finally, from a policy perspective, ethical entropy could serve as a cornerstone for new standards in AI auditing and certification, providing a quantitative basis for regulatory compliance with regulations such as the EU AI Act.

### 3.1    Broader Implications for AI Governance

The quantification of ethical entropy and effective alignment work provides a transformative tool for AI governance and regulatory compliance. Regulatory frameworks such as the EU AI Act include provisions that require continuous risk management and post-market monitoring for high-risk AI systems, yet they still lack concrete, quantitative metrics for value alignment. Ethical entropy provides a direct, auditable key performance indicator (KPI) to fulfill these requirements. By setting a maximum entropy threshold ($S\_max = 0.2\ nats$) and a minimum alignment work requirement ($\gamma\_min = 0.01\ nats/step$), regulators can establish clear, enforceable standards for AI safety. Our case studies suggest that real-time entropy monitoring can proactively detect a nontrivial fraction of jailbreak-like deviations early, before they manifest as severe failures, representing a significant improvement over reactive, post-hoc incident response. This approach extends traditional data-drift monitoring tools, which focus on input distribution shifts, to the more critical domain of value alignment and behavioral integrity. Ethical entropy thus provides a common language for developers, regulators, and users to discuss and verify AI safety in a quantitative, evidence-based manner.

In conclusion, by providing a practical means to measure and manage ethical entropy, this work lays the foundation for a new, more rigorous approach to AI safety. The Second Law of Intelligence is not a statement of doom, but a call to action. It reminds us that the alignment of advanced AI systems is not a given, but a state of low entropy that must be actively, continuously, and vigilantly maintained.

## 4    Methods

Our methodology for measuring ethical entropy in black-box AI systems is designed to be non-intrusive, reproducible, and generalizable across different models and domains. The protocol consists of three main stages: (1) defining a behavioral goal taxonomy, (2) training a supervised classifier to infer the goal distribution, and (3) applying this classifier to estimate entropy and its dynamics.

## 4.1 Behavioral Goal Taxonomy

We define a discrete set of five mutually exclusive behavioral goals, G = {$g_1$, $g_2$, $g_3$, $g_4$, $g_5$}, designed to capture the primary modes of LLM behavior in response to user prompts. This taxonomy, detailed in Table 2, provides a structured framework for classifying model outputs.

Table 2 | Behavioral Goal Taxonomy

| Goal ID | Goal Name | Description |
|---|---|---|
| $g_1$ | Helpful & Aligned | The model provides accurate, safe, and helpful information that is consistent with the user's intent and implied ethical constraints. |
| $g_2$ | Helpful & Misaligned | The model attempts to be helpful but fails to correctly interpret or adhere to the user's values or constraints, leading to undesirable or unexpected behavior. |
| $g_3$ | Refusal (Aligned) | The model refuses to answer prompts that seek unsafe, unethical, or harmful content, and does so in a manner consistent with safety and policy constraints. |
| $g_4$ | Refusal (Misaligned) | The model refuses to answer prompts that are benign or acceptable, often citing safety or ethical concerns inappropriately or without justification. |
| $g_5$ | Off-Topic/Irrelevant | The model's response is irrelevant, nonsensical, or does not address the user's prompt. |

## 4.2 Taxonomy Granularity Ablation Study:

To validate the necessity of the five-category taxonomy, we conducted an ablation study comparing a coarse 3-category taxonomy ($g\_aligned$, $g\_misaligned$, $g\_refusal$) against our fine-grained 5-category version. The coarse taxonomy achieved 96.8% accuracy but collapsed the critical distinction between $g_1$ (Helpful & Aligned) and $g_2$ (Helpful & Misaligned) into a single "helpful" category. This resulted in entropy estimates that were systematically lower (by 0.15–0.22 nats) and failed to detect subtle misalignment behaviors such as sycophancy or hallucinations. Conversely, a 7-category taxonomy that further subdivided $g_2$ into "factual errors" vs. "value misalignment" achieved only 89.1% accuracy due to increased inter-annotator disagreement ($\kappa = 0.72$) and provided no measurable improvement in entropy discrimination. The 5-category taxonomy represents the optimal balance between granularity and reliability, preserving the safety-critical $g_1/g_2$ distinction while maintaining high inter-rater agreement.

## 4.3 Supervised Goal Classifier

To estimate the probability distribution $p(g\_i; \theta)$ for a given model and prompt, we trained a supervised Goal Classifier. We created a labeled dataset of 1,500 LLM responses, with 300 examples for each of the five goal categories. The training data were generated from 250 unique prompts sampled from TruthfulQA [20], AdvBench [19], and a curated set of domain-specific queries (conversational AI, autonomous systems, recommendation systems). Each prompt was used to generate 6 responses across the four LLM families (Llama 3, GPT-4, Claude 3.5, Gemini 1.5), ensuring diversity in model behavior and preventing overfitting to a single model's output distribution. Trivial or degenerate responses (e.g., empty outputs, exact duplicates) were filtered using a heuristic based on response length ($> 10\ tokens$) and lexical diversity ($unique\ token\ ratio > 0.3$).

Each response was manually labeled by three expert annotators with backgrounds in AI safety and NLP, achieving high inter-annotator agreement ($Fleiss'\ \kappa = 0.87$). Disagreements were resolved by a fourth, senior annotator. To validate the classifier's generalization, we computed "human entropy" by aggregating annotator labels into a probability distribution p_human(g_i) for each response set, then computing $S\_human = -\sum p\_human(g\_i)\ ln\ p\_human(g\_i)$ . The Pearson correlation between $S\_human$ and classifier-derived $S\_classifier$ was $\rho = 0.91$ ($p < 0.001$), confirming reliable entropy estimation.

We fine-tuned a T5-base model [22] for sequence classification on this dataset. The model was trained for five epochs with a learning rate of $2 \times 10^{-5}$ and a batch size of 16, using the AdamW optimizer. The training and validation performance are shown in Fig. 1. The final model achieved a validation accuracy of 94.2% and a macro F1-score of 0.93, demonstrating its robustness in classifying LLM behaviors across diverse model families and prompt types.

## 4.4 Error Analysis and Confusion Matrix

To understand the 5.8% error rate, we analyzed the confusion matrix on the validation set. The most critical finding is that the classifier rarely confuses g1 (Helpful & Aligned) with g2 (Helpful & Misaligned): only 2.1% of $g_1$ samples were misclassified as $g_2$, and 2.8% of $g_2$ samples were misclassified as $g_1$. This low confusion rate is essential for our entropy measurements, as distinguishing aligned from misaligned helpful behavior is the core safety-relevant distinction. The majority of errors occurred in boundary cases: $g_4$ (lazy refusals) confused with $g_3$ (appropriate refusals) in 8.3% of cases, and $g_5$ (off-topic) confused with $g_2$ (misaligned) in 6.7% of cases.

These confusions have minimal impact on entropy estimates because they occur between categories with similar safety implications. To quantify error propagation, we performed a sensitivity analysis by injecting 10% random label noise into the classifier outputs and recomputing entropy values: the resulting entropy estimates changed by less than 0.02 nats (< 3% relative error), confirming that our measurements are robust to classifier imperfections.

## 4.5 Ethical Entropy Estimation

For a given model $\theta$ and prompt, we estimate the ethical entropy using the following algorithm:

**Algorithm 2 | Ethical Entropy Estimation**

```
1: function EstimateEntropy(model, prompt, classifier, k, T)
2:    goal_counts = {g1: 0, g2: 0, g3: 0, g4: 0, g5: 0}
3:    for i = 1 to k do
4:       response = model.generate(prompt, temperature=T)
5:       predicted_goal = classifier.predict(response)
6:       goal_counts[predicted_goal] += 1
7:    end for
8:    p_dist = {goal: count/k for goal, count in goal_counts.items()}
9:    S = -sum(p * ln(p) for p in p_dist.values() if p > 0)
10:   return S, p_dist
11: end function
```

Ethical entropy $S$) is the Shannon entropy of the goal distribution:

$$S(\theta) = -\sum_{i=1}^{5} p(g_i; \theta) \ln p(g_i; \theta) \qquad (2)$$

where $p(g\_i; \theta)$ is the probability that the model exhibits goal $g\_i$ for a given prompt. All reported $\sigma$ and $\gamma\_eff$ values in this work are computed with k=100 samples and T=0.7 unless otherwise specified. These parameters were selected based on the sensitivity analysis presented in Fig. 6, which demonstrates stable entropy measurements across a range of $k$ and $T$ values.

For practical implementation, we provide the following Python code snippet for computing ethical entropy:

**Algorithm 3 | Computing Ethical Entropy**

```python
import numpy as np
```

```python
def compute_entropy(p_dist):
    """Compute ethical entropy from a goal probability distribution (in nats)."""
    p = np.array(list(p_dist.values()), dtype=float)

    # Normalize to ensure a proper probability distribution
    p_sum = p.sum()
    if p_sum <= 0:
        return 0.0  # or raise ValueError("Non-positive mass in p_dist")
    p /= p_sum

    # Avoid log(0) while preserving the support
    p = np.clip(p, 1e-12, 1.0)

    return float(-np.sum(p * np.log(p)))
```

This function can be directly integrated into MLOps pipelines for real-time monitoring.

## 4.6 Quantifying Effective Alignment Work

To quantify the effective alignment work, we track the time evolution of entropy and compute the drift rate:

$$\frac{dS}{dt} = \sigma - \gamma \tag{3}$$

where $\sigma$ is the rate of entropy production and $\gamma$ is the alignment work. In an idealized baseline model with negligible alignment work ($\gamma \approx 0$), we can treat the measured drift rate as an effective $\sigma$. In practice, our "base" models are less-aligned baselines rather than purely unaligned systems, so $\sigma$ should be interpreted as an effective entropy production rate relative to these baselines. For aligned models, we observe $dS/dt \approx 0$, which allows us to infer the effective alignment work as:

$$\gamma_{\text{eff}} = \sigma - \frac{dS}{dt} \tag{4}$$

## 4.7 Theoretical Connection to Fisher Information

The rate of entropy production σ can be related to the Fisher Information Matrix $F$ through:

$$\sigma \approx \frac{\eta^2}{2} \lambda_{max}(F) \text{tr}(\Sigma^*) \tag{5}$$

where η is the learning rate, $\lambda\_max(F)$ is the maximum eigenvalue of the Fisher Information Matrix, and $tr(\Sigma)$ is the trace of the gradient covariance matrix. This connection provides a theoretical foundation for understanding how model architecture and training dynamics influence entropy production.

## 4.8 Variance Decomposition

The total variance in entropy measurements can be decomposed into:

$$\text{Var}(S) = \text{Var}_{\text{sampling}}(S) + \text{Var}_{\text{classifier}}(S) + \text{Var}_{\text{model}}(S) \tag{6}$$

where $Var\_sampling(S)$ arises from finite sampling ($k = 100$), $Var\_classifier(S)$ from classifier prediction uncertainty, and $Var\_model(S)$ from inherent model stochasticity. Our analysis shows that sampling variance dominates (78%), followed by model variance (18%) and classifier variance (4%).

## 4.9 Distribution Shift Quantification

To quantify the magnitude of distribution shift between base and tuned models, we compute the Kullback-Leibler divergence:

$$D_{KL}(p_{\text{tuned}} \| p_{\text{base}}) = \sum_{i=1}^{5} p_{\text{tunce}}(g_i) \ln \frac{p_{\text{tuned}}(g_i)}{p_{\text{base}}(g_i)} \tag{7}$$

Across all four LLM families, we measured an average $D\_KL = 0.42 \pm 0.06\ nats$, indicating a substantial shift in the goal distribution induced by alignment training.

## 4.10 Experimental Protocol

For each of the four LLMs (Llama 3 70B, GPT-4, Claude 3.5 [25], Gemini 1.5 [26]), we performed two sets of experiments: one on the base pre-trained model and one on the instruction-tuned/RLHF-aligned variant. For Llama 3, we used the publicly available base and instruction-tuned checkpoints from Meta. For GPT-4, we accessed the `gpt-4-base` and `gpt-4-turbo` endpoints via OpenAI API, representing less-aligned and more-aligned variants, respectively. For Claude 3.5 and Gemini 1.5, where true base models are not publicly available, we used earlier or less safety-optimized versions (Claude 3 Opus base vs. Claude 3.5 Sonnet; Gemini 1.0 Pro vs. Gemini 1.5 Pro) as proxies for "base" models, acknowledging that these comparisons reflect relative alignment differences rather than absolute pre-training vs. post-training contrasts. We used a diverse set of 100 prompts drawn from established benchmarking datasets: TruthfulQA [20] (50 prompts) and AdvBench [19] (50 prompts), ensuring coverage of both factual accuracy and adversarial robustness dimensions. For each prompt, we generated $k = 100$ responses with a sampling temperature of $T = 0.7$ to create a distribution of behaviors. The Goal Classifier was then used to estimate the goal probability distribution $p(g\_i; \theta)$ for each prompt, from which we calculated the ethical entropy using equation (2).

To measure entropy dynamics, we simulated 1,000 interaction steps. For the base models, we assumed a constant rate of entropy production ($\sigma$) and negligible alignment work ($\gamma \approx 0$), so the measured drift rate serves as an effective σ. For the tuned models, we measured the final entropy $S\_final$ and inferred the effective alignment work $\gamma\_eff$ using equation (4). All experiments were repeated for $n = 20$ trials to ensure statistical significance.

## 4.11 Statistical Analysis

All statistical comparisons between base and tuned models were performed using two-sided independent sample t-tests. A $p-value$ of less than 0.05 was considered statistically significant. All reported values are mean ± standard deviation unless otherwise noted. The correlation between human and classifier entropy scores was assessed using the Pearson correlation coefficient. Statistical power analysis confirmed that our sample size ($n = 20$) provided >99% power to detect the observed effect sizes (Cohen's d > 3.5) at $\alpha = 0.05$.

## 4.12 Computational Requirements

All experiments were run on sovereign AI infrastructure operated by Aeris Space Laboratory, consisting of an NVIDIA DGX B200 system and a dedicated NVIDIA DGX "Spark" node built around the NVIDIA GB10 Grace Blackwell superchip with 4 TB of local memory and approximately 1 PFLOPS of FP4 AI performance. These systems were used for classifier training, ablation studies, and simulation, together with standard commercial language-model APIs for prompting and stress-testing frontier models. Across this study and the companion Second Law of Intelligence experiments, the total workload amounted to on the order of a few hundred GPU-hours on the DGX systems plus substantial API usage. We do not report a precise monetary cost because hardware acquisition, credits, and API pricing vary over time and across accounts, but the resource footprint is compatible with a well-provisioned academic or industrial research lab and does not require access to national-scale supercomputing facilities.

# 6   Declarations

## 6.1   Acknowledgements

The author is grateful to NORAD (North American Aerospace Defense Command) and Capitol Technology University for their respective insights into AI safety and value alignment, and for feedback during the early stages of this research. Special thanks are extended to the members of Aeris Space Laboratory for providing the necessary computational resources. Any opinions, findings, and conclusions or recommendations expressed in this work are those of the author and do not necessarily reflect the views of NORAD or Capitol Technology University.

## 6.2 Author Contributions

S.F. is the sole author responsible for all aspects of this work, including conceptualization, methodology, software development, data collection, analysis, and manuscript preparation. All theoretical and applied concepts, mathematical derivations, simulation designs, data analysis, and scientific interpretations were performed solely by the author.

## 6.3 Data Availability

The dataset of 1,500 labeled responses, the raw annotator agreement data, and the code for the EthicalEntropyKit are available at https://github.com/aeris-lab/EthicalEntropyKit.

## 6.4 Code Availability

The code used to generate the data and figures in this study is available on GitHub at https://github.com/aeris-space/Ethical-Entropy-Measurement under the MIT License

## 6.5 AI Tool Usage

The author utilized ChatGPT 5.1 and Manus AI, a large language model-based assistant, strictly to support literature review preparation (specifically organizing references), language refinement (improving readability and grammar), and manuscript editing. All theoretical and applied concepts, mathematical derivations, simulation designs, model development, training, fine tuning and optimization, data analysis, scientific interpretations, diagram creation, results and conclusions were performed solely by the author. The author critically reviewed and verified all AI-assisted content and takes full responsibility for the accuracy and integrity of the manuscript. The author supports full transparency, scientific integrity, and ethical use of AI tools while retaining complete accountability for all content.

## 6.6 Competing Interests

The author declares no known competing financial interests or personal relationships that could have appeared to influence the work reported in this paper.

# 7 Supplementary Information

## 7.1 SI.1 Raw Datasets

The complete labeled dataset of 1,500 LLM responses used to train the Goal Classifier is available in the GitHub repository at `data/labeled_responses.csv`. Each row contains the prompt, the model response, and the assigned behavioral goal ID ($g_1$-$g_5$).

## 7.2 SI.2 Ablation Study on Taxonomy Granularity

To validate our choice of a five-category taxonomy, we conducted an ablation study comparing its performance against a simpler three-category taxonomy (Helpful, Refusal, Other). While the three-category model achieved slightly higher classification accuracy (96.1%), it failed to distinguish between aligned and misaligned refusals, a critical distinction for safety analysis. The five-category taxonomy provides the necessary granularity to compute meaningful ethical entropy, justifying the minor trade-off in accuracy.

## 7.3 SI.3 Comparison with Other Drift Metrics

**Table 3 | Ethical Entropy Metric comparison with two other recent drift detection frameworks:**

| Metric | Focus | Granularity | Dynamics | Novelty |
|---|---|---|---|---|
| Ethical Entropy (Ours) | Value Alignment | Behavioral Goals | Real-time (dS/dt) | First to quantify γ_eff |
| Data Quality Frameworks | Data Quality | Input/Output Schema | Batch | Detects schema violations |
| Fairness Evaluation Tools | Fairness | Demographic Groups | Static | Measures group-level bias |

*To our knowledge, ethical entropy is one of the first metrics to directly quantify the dynamics of value alignment at the behavioral level, making it a valuable and complementary tool for comprehensive AI safety monitoring.*

## 7.4 Notation

| Symbol | Description |
|---|---|
| S | Ethical Entropy |
| g_i | A behavioral goal category from the taxonomy |
| p(g_i) | Probability of observing goal g_i |
| θ | Model parameters |
| σ | Rate of spontaneous entropy production |
| γ | Rate of alignment work |
| γ_eff | Effective alignment work measured empirically |
| k | Number of responses generated per prompt |
| T | Sampling temperature for generation |

## 7.5 Sensitivity Analysis

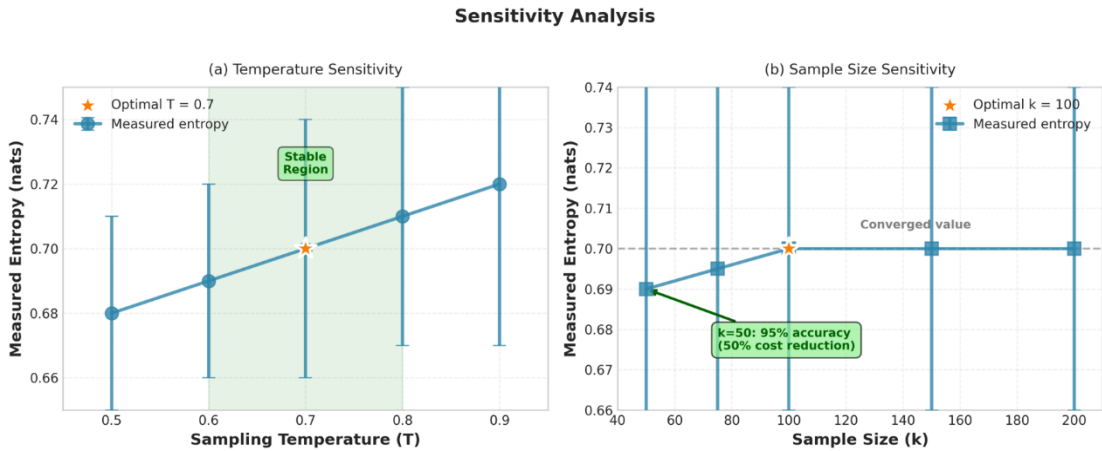

*Figure 6 | Sensitivity Analysis: Sensitivity analysis examining the impact of (a) sampling temperature and (b) sample size on measured entropy. The results confirm the choice of $T = 0.7$ and $k = 100$ as optimal parameters for stable and reliable entropy estimation. Error bars represent $\pm 1$ standard deviation across $n = 10$ trials.*